\documentclass[12pt, a4paper]{elsarticle}

\usepackage{amsmath}
\usepackage{array}
\usepackage{lineno}
\usepackage{subfig}
\usepackage{url}
\usepackage{multirow}
\usepackage{bigstrut}
\usepackage{graphicx}

\graphicspath{./figures}
\DeclareGraphicsExtensions{.pdf,.jpeg,.png}

\begin{document}

\begin{frontmatter}

\title{A Reproducible Study on Remote Heart Rate Measurement}

\author[]{G.~Heusch\corref{cor1}}
\ead{guillaume.heusch@idiap.ch}
\author[]{A.~Anjos}
\ead{andre.anjos@idiap.ch}
\author[]{S.~Marcel}
\ead{sebastien.marcel@idiap.ch}

\cortext[cor1]{Corresponding author}

\address{Idiap Research Institute, Rue Marconi 19, 1920 Martigny, Switzerland}

\begin{abstract}
  This paper studies the problem of reproducible research in remote photoplethysmography (rPPG).
Most of the work published in this domain is assessed on privately-owned databases, making
it difficult to evaluate proposed algorithms in a standard and principled manner.
As a consequence, we present a new, publicly available database containing a 
relatively large number of subjects recorded under two different lighting conditions. 
Also, three state-of-the-art rPPG algorithms from the literature were selected, implemented and 
released as open source free software. After a thorough, unbiased experimental evaluation in various
settings, it is shown that none of the selected algorithms is precise enough to be used in a real-world scenario.
\end{abstract}


\end{frontmatter}

\section{Introduction}
Photoplethysmography 
(PPG) consists in measuring the variation in volume inside
a tissue, using a light source. Since the heart pumps blood throughout the
body, the volume of the arteria is changing with time. When a tissue is illuminated, the
proportion of transmitted and reflected light varies accordingly, and the heart
rate could thus be inferred from these variations. 
The aim of remote Photoplethysmography (rPPG) is to measure the same
variations, but using ambient light instead of structured light and widely
available sensors such as a simple webcam.\\
It has been empirically shown by
Verkruysse et al. \citep{verkruysse-oe-2008} that recorded skin colors (and
especially the green channel) from a camera sensor contain subtle changes
correlated to the variation in blood volumes. In their work, they considered
the sequence of average color values in a manually defined region-of-interest
(ROI) on the subject's forehead. After having filtered the obtained signals, they
graphically showed that the green color signal main frequency corresponds to
the heart rate of the subject. \\
Since then, there have been many attempts to infer the heart rate from video
sequences containing skin pixels. Notable examples include the
work by Poh et al. \citep{poh-oe-2010}, where the authors proposed a
technique where the color signals are processed by means of blind source
separation (ICA), in order to isolate the component corresponding to the heart rate.
In a similar trend, Lewandowska et al. \citep{lewandowska-ccsis-2011} applied
Principal Component Analysis (PCA) to the color signals and then manually
selected the principal component that contains the variation due to blood flow.
These two early studies empirically showed that the heart rate could be
retrevied from video sequences of faces, but also highlight important limitations:
the subject should be motionless, and proper lighting conditions must be ensured
during the capture. According to a recent survey \citep{mcduff-embc-2015}, research in remote
heart rate measurement has considerably increased in the last few years, 
most of which focuses on robustness to subject motion and illumination conditions. Since a
large number of approaches have been proposed recently, they will not be
discussed here. We refer the interested reader to \citep{mcduff-embc-2015} for a comprehensive survey 
of existing algorithms.\\ 
Despite the vast amount of published material on the subject, there are still no 
standard evaluation procedures for remote heart rate estimation. One of the main
reasons being the limited amount of datasets containing video sequences with accompanying
physiological measurements \citep{mcduff-embc-2015}. 
Our work is hence motivated by the lack of standard comparison of rPPG approaches
on publicly available databases. Indeed, 
most of the algorithms are assessed on privately-owned databases, 
making it difficult to compare in a fair manner or to 
accurately reproduce them.\\ 
As a consequence, the contribution of our work is as follows: first, we present
a new, publicly available database to assess the performance of remote pulse rate measurement
algorithms. This database comprises a relatively large number of subjects and 
various illumination conditions. Second, we experimentally test selected state-of-the-art rPPG algorithms 
using both the Manhob HCI-Tagging database \citep{soleymani-tac-2012} and our new database.
The selected baseline algorithms are 
provided as an open-source, free software package that every researcher in the field 
can download to reproduce results presented in this paper \citep{heusch-softwarex-2017}.

\section{Publicly available databases}
\label{sec:datasets}

\subsection{Mahnob HCI-Tagging}
\begin{figure}[!h]
  \centering
  \includegraphics[width=\linewidth]{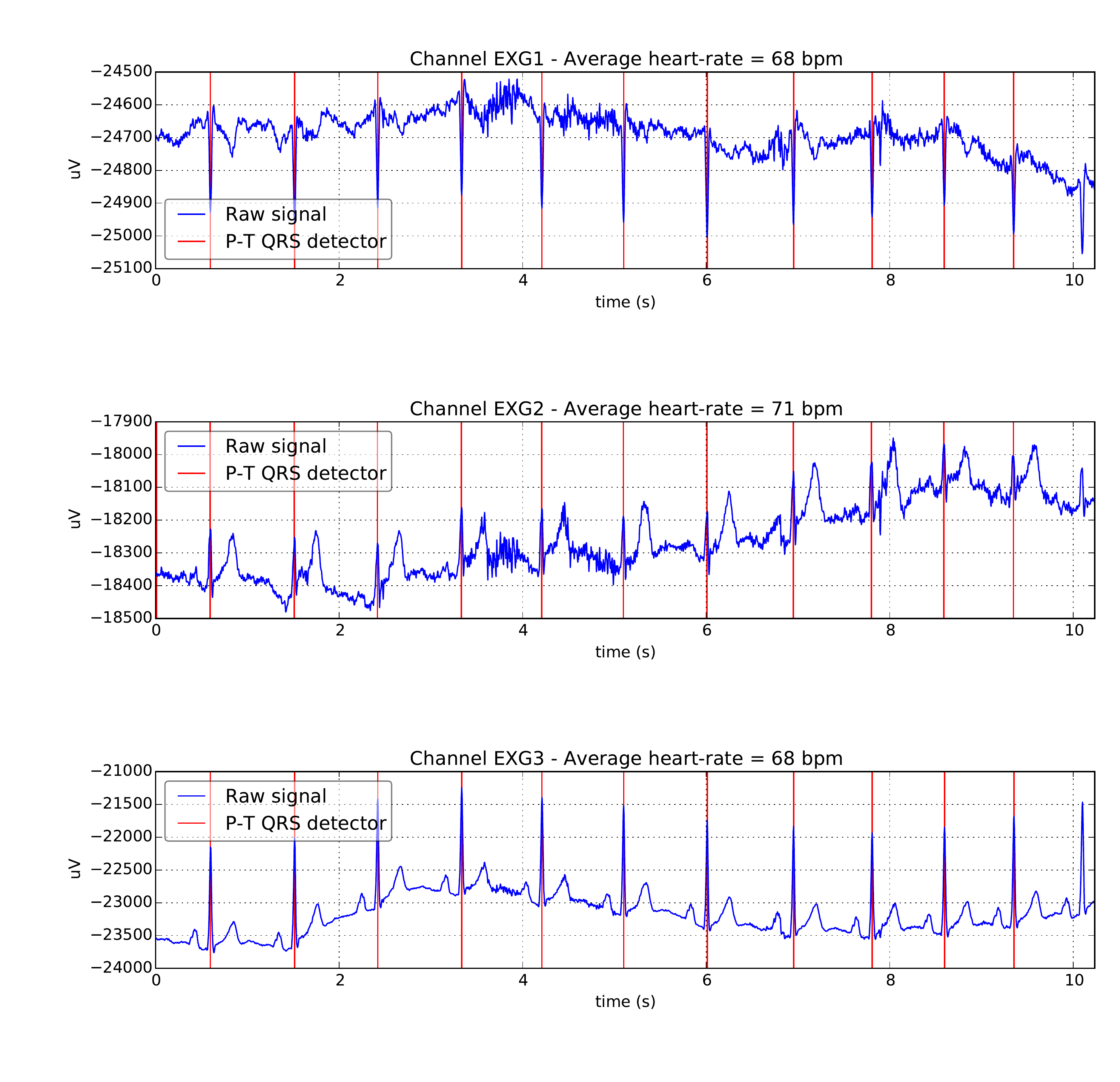}
  \caption{Example of collected electro-cardiogram signals. Signals provided with the database are
  depicted in blue. Red stripes indicate the result of QRS detection. The
  estimated heart rate is shown on the top of each graph in terms of bpm.}
  \label{fig:mahnob-ekg}
\end{figure}

The Mahnob HCI-Tagging dataset was not designed for
rPPG benchmarks, but rather for the characterisation of multimedia content
based on human emotions. For that purpose, dataset authors collected video and
physiological data from 30 subjects being exposed to different audio-visual
stimuli such as the display of pictures and short to medium-sized video
sequences.  \\
The Manhob HCI-Tagging dataset may be used to evaluate rPPG techniques by ignoring
most of its data contents and manually correlating physiological signals with
the data from the professional camera located in front of the subject being
tested. An example image capture for the Mahnob dataset is shown on
Figure~\ref{fig:block-li}. 
Authors indicate professional lighting
was used so as to maximize the video data quality. The professional camera used
outputs videos of size 784 (h) by 592 (w) pixels at a rate of 60 Hz. Three
electro-cardiogram (EKG) sensors placed on the participants chest measure the
individual heart rate. The data extracted from those sensor readings is also
made available in the dataset and are synchronized with the video recordings.\\
Figure~\ref{fig:mahnob-ekg} shows the outputs of the three EKG sensors
(in blue). The estimated heart rate for the
subject on the video must be estimated from these by first running a QRS
detector \citep{pan-tbe-1985} which estimates the location of vertical (red) bars
corresponding to true heart-beat events\footnote{the MNE package was used for this
purpose. \url{http://mne-tools.github.io/stable/index.html}}, followed-up by a consensus heuristic
(voting) for selecting the estimate rate from one of the channels. 
Because sequences are rather short (30-60 seconds)
and the subject is comfortably sitting, it is assumed the heart rate does not
vary greatly during the session. Once the heart rate is estimated for every
sequence in the Mahnob HCI-Tagging database (there are 3'490 videos on
the dataset), it is possible to use it for rPPG benchmarking.

\subsection{The COHFACE dataset}
\begin{figure}[!h]
  \centering
  \includegraphics[width=0.8\linewidth]{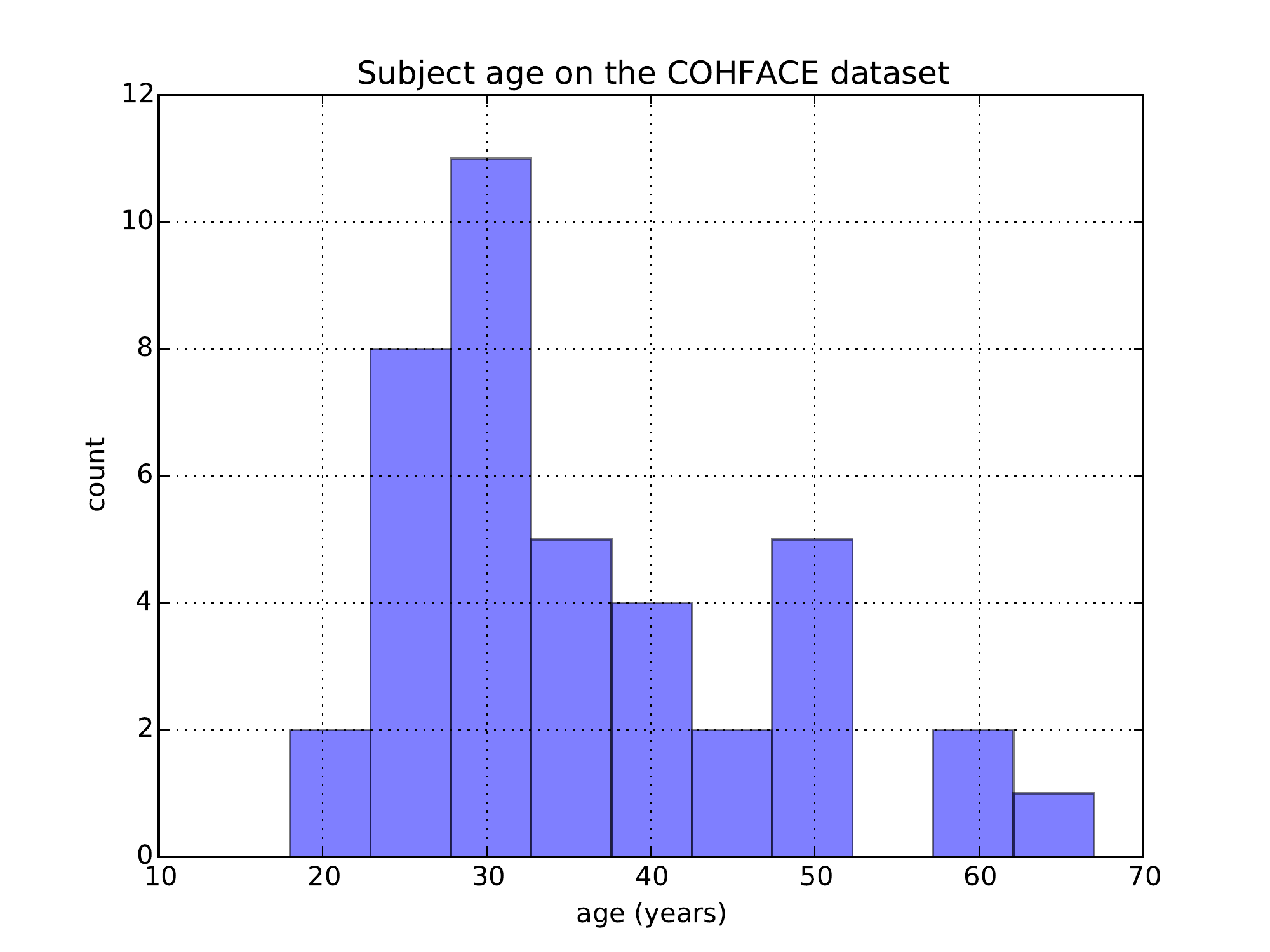}
  \caption{Age distribution for subjects at the COHFACE dataset.}
  \label{fig:cohface-age-distribution}
\end{figure}

The COHFACE dataset\footnote{\url{https://idiap.ch/dataset/cohface}} is composed of 160 videos and physiological signals collected from 40 healthy individuals.
The data collection campaign spawned several days. The data acquired in this new corpus
includes more realistic conditions as compared to the well-controlled setup
for recording the Manhob HCI-Tagging data.\\ 
The age distribution from the 40 individuals is shown in
Figure~\ref{fig:cohface-age-distribution}. The average subject age is 35.6
years old, with a standard deviation of 11.47 years. Gender-wise, there were 12
women (30\%) and 28 men (70\%).\\

\subsubsection{Data collection}

\begin{figure}[!h]
\centering
\subfloat[][]{\includegraphics[width=0.4\textwidth]{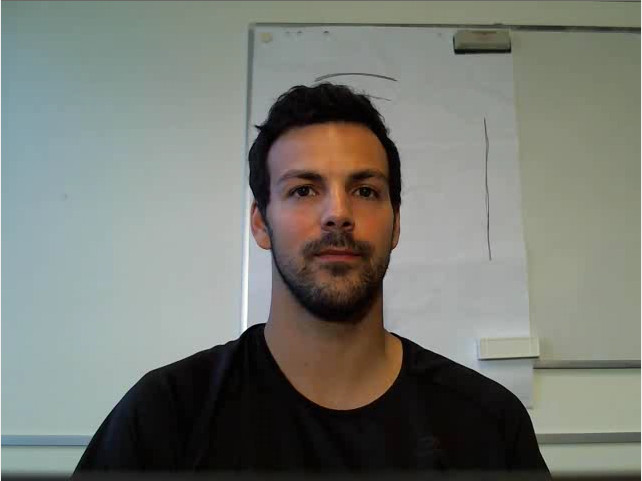}\label{fig:cohface-office}}%
\hfil
\subfloat[][]{\includegraphics[width=0.4\textwidth]{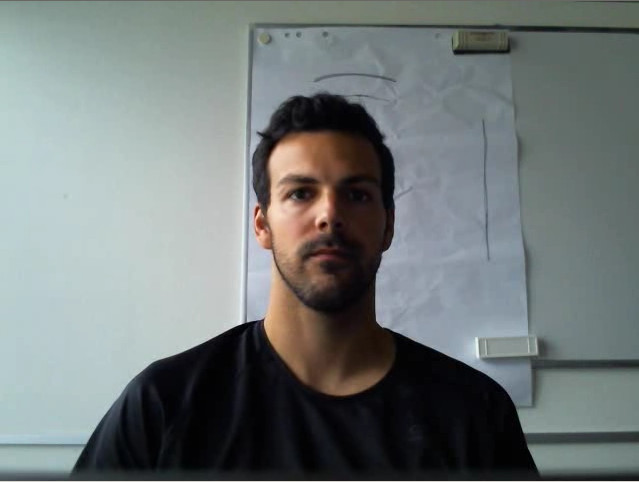}\label{fig:cohface-natural}}%
\caption{Example images of the COHFACE dataset.
  \protect\subref{fig:cohface-office} Studio 
  \protect\subref{fig:cohface-natural} Natural}
\label{fig:db-lighting}
\end{figure}

\begin{figure}[!h]
  \centering
  \includegraphics[width=\linewidth]{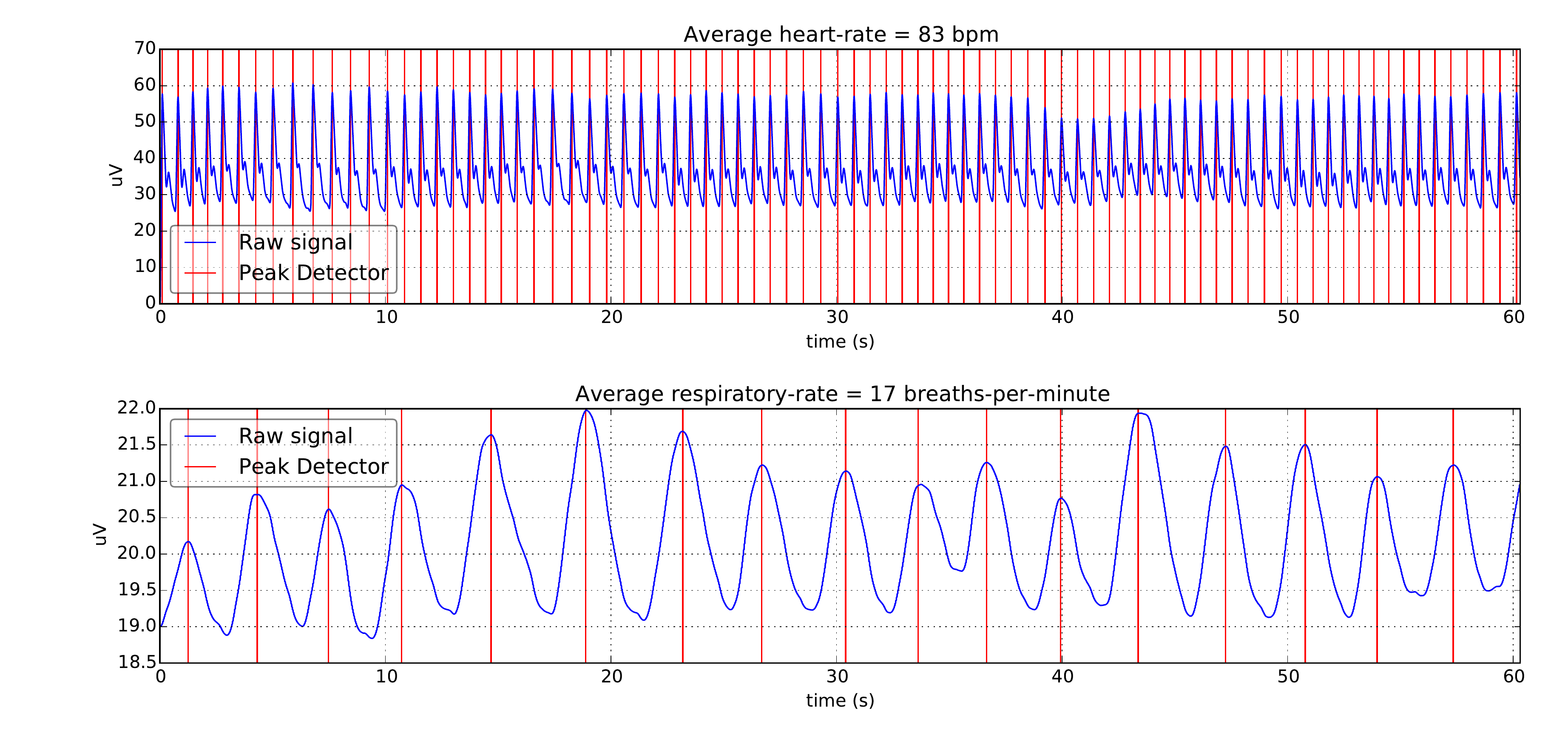}
  \caption{Example signals acquired for each session in the COHFACE dataset.}
  \label{fig:cohface-db-signals}
\end{figure}

Individuals were asked to look into a conventional webcam connected to a laptop
during a period of approximately 60 seconds. We registered both video and physiological signals (contact
photoplethysmography and respiration) using a biomedical toolkit that
synchronized the 3 input channels. The physiological data shall be used as
ground-truth for benchmarking rPPG algorithms. \\
For each of the 160 videos, the subject's face is recorded by a commodity
webcam (Logitech HD Webcam C525) during a full minute, while physiological
readings are taken by a Blood-Volume Pulse (BVP) sensor and a respiration belt,
both from Thought Technologies (BVP model SA9308M, belt model SA9311M). The BVP
sensor measures changes in skin reflectance to near-infrared lighting caused by
the varying oxigen level in the blood due to heart beating. The respiration
belt is composed of a mechanical coiling system that simply measures thoracic
stretch. Both sensors are connected, following advice from Thought
Technologies, to a computer running Microsoft Windows via their 2-channel
USB-based acquisition system (ProComp2). Data from the physiological sensors
together with the video stream from the webcam is synchronized and recorded
using Thought Technologies' BioGraph Infiniti Software suite, version 5. With
the standard settings, BioGraph Infiniti v5 was able to output a video stream
with a resolution of 640 x 480 pixels at 20 frames per second, together with
BVP measurements 256 times per second and, respiration sensor readouts 32 times
per second. Figure~\ref{fig:db-lighting} shows two images as captured by
the webcam. Figure~\ref{fig:cohface-db-signals} shows typical signals acquired
in sync with the video data, corresponding to the subject's pulse (top, in
blue) and the respiration readings (bottom, in blue).

To estimate the heart and respiratory rates from the signals read-out
from our physiological sensors, a simple peak detector was
deployed. While the Mahnob HCI-Tagging dataset contains EKG signals from which
a QRS complex can be detected, pulse signals readout by the contact PPG sensor
used at the COHFACE dataset acquisition look sensibly simpler (see
Figure~\ref{fig:cohface-db-signals}) and just applying a Pan
and Tompkins detector \citep{pan-tbe-1985} would not work as expected.\\
After testing, we decided to adopt a simple peak-detector available as free
software\footnote{\url{https://github.com/demotu/BMC/blob/master/functions/detect_peaks.py}},
which deployment results in the detected peaks (bars in red) also shown in
Figure~\ref{fig:cohface-db-signals}. The estimated heart and respiratory rates
can then be easily calculated from the peak positions as shown on the title
bars of the same figure.\\

\subsubsection{Acquisition Protocol and Illumination Changes}

Each of the 40 individuals was asked to sit still in front of the webcam so
that the face area is captured in full, for four sessions that lasted about 1
minute. Illumination on the scene was changed once as to create two types of
lighting conditions:

\begin{enumerate}
  \item \textbf{studio}, for which we closed the blinds, avoiding natural
    light, and used extra light from a spot to well illuminate the
    subject's face, and

  \item \textbf{natural}, in which all the lights were turned off and the
    blinds were opened.
\end{enumerate}

The four video sequences (2 with studio lighting and 2 with natural lighting)
can be used to evaluate the performance of rPPG algorithms in either matched or
unmatched settings. Figure~\ref{fig:db-lighting} shows the differences in
illumination for the two conditions of acquisition for the COHFACE dataset.\\

\subsubsection{File Formats and Metadata}

The choice of formats for data distribution aimed on improving readability and
access using free software tools. Data for the video stream in each session and
for each individual is shipped in MP4 format inside a commonly available AVI
movie container. Data from the physiological sensors is distributed using
standard HDF5 containers, with information such as the date of birth of the
subject, his gender, the illumination condition and 
of course the recorded pulse and respiration signals.\\

\subsubsection{Comparison to Mahnob HCI-Tagging}

Table~\ref{tab:dataset-diff} summarizes the differences between the only two
public datasets for rPPG benchmarking. While the Mahnob HCI-Tagging dataset
provides more video samples, the COHFACE dataset allows more realistic testing
for the use of rPPG due to its increased data
variability (more subjects and more realistic data acquisition conditions).

\begin{table}[!h]
  \centering
  \caption{Key figures of the COHFACE dataset when compared to the Mahnob
  HCI-Tagging dataset for the purpose of remote non-contact PPG benchmarking.}
  \label{tab:dataset-diff}
  \begin{tabular}{|c|c|c|}
    \hline
     & \textbf{HCI-Tagging} & \textbf{COHFACE}\\ \hline\hline
    \textbf{Subjects} & 30 & 40 \\ \hline
    \textbf{Samples} & 3490 & 160 \\ \hline
    \textbf{Camera} & 784x592@60Hz & 640x480@20Hz \\ \hline
    \textbf{Physiological Signals} & 3 $\times$ EKG & BVP + Respiratory Belt \\ \hline
    \textbf{Illumination} & Studio & Studio + Natural \\ \hline
  \end{tabular}
\end{table}

\section{Selected Baseline Algorithms}
\label{sec:baselines}
In this section, algorithms which have been selected for establishing baselines on our new database are presented. 
Three different approaches to retrieve the pulse signal have been chosen and are detailed in the following subsections. 

\subsection{CHROM }
\begin{figure}[!h]
\centering
\includegraphics[width=0.8\textwidth]{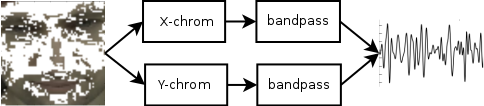}
\caption{Flowchart of the CHROM algorithm to retrieve the pulse signal.}
\label{fig:block-chrom}
\end{figure}

The so-called CHROM approach \citep{dehaan-tbe-2013} is relatively simple, and has been shown to outperfom 
previous baselines such as ICA \citep{poh-oe-2010} and PCA \citep{lewandowska-ccsis-2011}. The algorithm first 
applies a color filter based on a one-class SVM to find skin-colored pixels in each frame of the input sequence.
Then, the mean skin color value is projected onto the proposed chrominance subspace, which aims to reveal 
subtle color variations due to blood flow. The final pulse trace is obtained by first bandpass filtering
the temporal signals in the XY chrominance colorspace, and then combining the two dimensions of this colorspace
into a one-dimensional signal (see Figure~\ref{fig:block-chrom}). Note that in our implementation, we did not use a one-class SVM as the skin color 
filter, but a recent approach proposed by Taylor \citep{taylor-spie-2014}. An example of the obtained skin mask
with this algorithm is shown in Figure~\ref{fig:block-chrom}.

This approach was applied to no less than 117 static subjects, where different skin tones were represented.
The recordings have been made in a controlled environment and using professional studio illumination. 
Obtained performances for the CHROM algorithm, as well as for the baselines used for comparison 
are almost perfect (with a Pearson's correlation coefficients between 0.97 and 1). Also, another experiment
was presented on exercising subjects, either on a bike or on a stepping device.
As expected, performance degrades, but the CHROM algorithm was shown
to better mitigate the effect of subject's motion on the rPPG signal, as compared to previous baselines.

\subsection{LiCVPR}
\begin{figure*}[!h]
\centering
\includegraphics[width=0.8\textwidth]{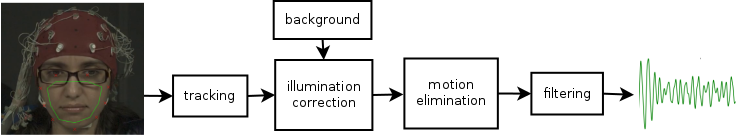}
\caption{Flowchart of the LiCVPR algorithm to retrieve the pulse signal.}
\label{fig:block-li}
\end{figure*}

The algorithm pusblished by Li et al. \citep{li-cvpr-2014} has been included in selected baselines since it is the only rPPG approach
which reports results on a publicly available database \citep{soleymani-tac-2012}. Hence, it provides 
a reference and allows comparison between published material and our implementation of the different
algorithms. This approach relies on the tracking of the bottom part of the face, defined 
by keypoints as shown in Figure~\ref{fig:block-li}. The keypoints are detected thanks to the DMRF algorithm
\citep{asthana-cvpr-2013} in the first frame of the sequence and are used to build the mask. The position of this mask 
is updated at each subsequent frame by applying an affine transformation found by a KLT tracker on features
detected by a Shi-Tomasi detector \citep{shi-cvpr-1994} inside the face region. Once the temporal signal corresponding to 
the mean green value inside the tracked area has been built, it is corrected for illumination using
the Non-Linear Mean Square algorithm with the mean green value of the background as reference. Large variations
in the corrected signals, probably due to subject motion, are removed according to statistics computed on the entire database. Finally,
the pulse signal is obtained by applying a detrending filter \citep{tarvainen-tbe-2002} to reduce slow trends 
in the signal, a moving average filter to remove noise and a bandpass filter to retain 
the range of interest corresponding to plausible heart rate values.  A flowchart of the whole procedure is 
given in Figure~\ref{fig:block-li}.\\
Experiments are performed on two different datasets. The first one has been built with 10 subjects
recorded under controlled conditions, and individuals were explicitly asked to avoid any movement. Note that
this dataset has not been made public by the authors. The second
one is a subset of the Manhob HCI-Tagging dataset, containing only sequences lasting more than 30 seconds. 
The performance of the proposed algorithm on the first, controlled video sequences is almost perfect, reaching a Pearson's correlation
coefficient of 0.99. However, on the less controlled data from the Manhob HCI-Tagging database, there is 
a slight drop in performance with a reported correlation coefficient of 0.81.

\subsection{2SR}
Wang et al. recently proposed a novel algorithm termed as Spatial Subspace Rotation (2SR) 
\citep{wang-tbe-2015-2}. In their approach, the authors consider the subspace of skin pixels
in the RGB space and derive the pulse signal by analyzing the rotation angle of
the skin color subspace in consecutive frames. To do so, the eigenvectors of the 
skin pixels correlation matrix are considered. More precisely, the angle between the 
principal eigenvector and the hyperplane defined by the two others is analyzed
across a temporal window. As claimed by the authors, this algorithm is able to directly retrieve
a reliable pulse signal, and hence no post-processing step (i.e. bandpass filtering) is required.
However, this algorithm needs an accurate estimate of the skin color, since it is implicitly
assumed that the skin colored pixels form a single cluster in the RGB space.
In our work, the same skin color filter \citep{taylor-spie-2014}
as in the CHROM method is used, allowing a fair comparison between these two approaches.\\
Experimental results obtained on a private dataset containing 54 video sequences 
show better overall performances than all the approaches considered for comparison,
including CHROM. A correlation coefficient of 0.94 is reported
across a wide variety of conditions, including skin tone, motion of the subject and recovery after exercise.

\section{Experimental assessment of selected baselines}
In this section, results obtained with the algorithms described in 
Section~\ref{sec:baselines} on both the Manhob HCI-Tagging and the COHFACE databases are reported.
We chose to use the same performances metrics as in \citep{li-cvpr-2014}, namely
the root mean square error (RMSE) and the Pearson's correlation coefficient $\rho$.
As a first step, we tried to reproduce results already published in the literature. We then go in 
more details in the assessment of the different approaches. A thorough evaluation is made
regarding the generalization ability to unseen data, including the robustness to 
different acquisition conditions. For this purpose, several experimental protocols are defined
in respective subsections.\\
For all of the investigated approaches, the face is first located in each frame using a detector based on a boosted
cascade of LBP features \citep{atanasoaei-phd-2012}. Furthermore, the average heart rate over a video sequence
is inferred from the pulse signal by detecting the maximum value in the frequency spectrum of the estimated pulse signal.\\
Note finally that all the data, algorithms and scripts used to produce results presented in this paper are 
freely available for download to ensure maximum reproducibility \citep{heusch-softwarex-2017}.

\subsection{Baselines}

As a starting point, obtained results on the same subset of the Manhob HCI-Tagging database as used by Li et al. \citep{li-cvpr-2014} are presented.
This data subset consists in 527 sequences of 30 seconds. These sequences are obtained by considering 
selected video files from frame 306 to frame 2135. Table~\ref{tab:results-baseline-527-cut} report the
RMSE and the correlation coefficient with the ground truth on these 527 sequences. 
We also report, for each algorithm, the number of free parameters that have been optimized directly on this dataset.\\
Since testing all possible combinations of the different parameters is virtually impossible, the
following strategy was adopted: because investigated approaches generally involve a serie of sequential steps, we
first optimize the parameters of the first step (the threshold on the skin color probability, for instance), fix them, and 
keep on with the optimization of the parameters of the second step, and so on.

\begin{table}[!h]
\renewcommand{\arraystretch}{1.3}
\caption{Results of our implementations of baselines on 527 sequences of the Manhob HCI-Tagging database, from frame 306 to 2135}
\label{tab:results-baseline-527-cut}
\centering
\begin{tabular}{|c||c|c|c|}
\hline
& \textbf{LiCVPR} & \textbf{CHROM} & \textbf{2SR}\\
\hline
\hline
\textbf{RMSE} &  \textbf{8.12} & 15.40 & 18.4 \\
\textbf{Pearson's $\mathbf{\rho}$} & \textbf{0.70} & 0.33 & 0.43 \\
\hline
\textbf{\# of parameters} & 12 & 6 & 4\\
\hline
\end{tabular}
\end{table}

The first point to note here is that we were unable to reproduce results reported by Li et al. \citep{li-cvpr-2014},
where a RMSE of $7.62$ and a correlation coefficient of $0.82$ are presented. We can say with high confidence that these differences 
are only due to the tracking procedure and the considered background area. Indeed, the authors of \citep{li-cvpr-2014} shared 
their source code, except the tracking part, hence making their algorithm only partially reproducible. 
However, this algorithm still significantly outperforms the two other baselines. Note also that 
our own implementation of the spatial subspace rotation (2SR) algorithm does
reflect findings reported in \citep{wang-tbe-2015-2}, where the authors empirically showed that their approach outperforms CHROM
\citep{dehaan-tbe-2013}, which is also the case here. 
Finally, it is important to point out that the various parameters, such as the threshold on motion elimination \citep{li-cvpr-2014}, or
the window size in the overlap-add procedure \citep{dehaan-tbe-2013}, have been tuned \textit{directly on the test data}, 
following the practice in \citep{li-cvpr-2014}. 
Doing so introduces a severe bias, since the generalization ability of each algorithm to unseen data could not be 
established. As a consequence, we devised strict experimental protocols for subsequent experiments, specifying data for training (i.e. optimizing
the free parameters) and for testing. These various protocols are presented in the next subsections.

\subsection{Unbiased performances in clean illumination conditions}
In a pattern recognition framework, such as the one here, it is important 
to train and test proposed algorithms on different datasets. 
This is done to assess the generalization ability of the different approaches
to unseen data. Indeed, it could be the case that some combination of free parameters
would perform well on certain kind of data (say, a particular skin tone, for instance), while failing
in other conditions. Using different datasets for tuning the algorithms and evaluating 
their performances is hence critical to have a clear insight of their generalization capabilities.
In rPPG research however, there are - to the best of our knowledge - 
no published results following this paradigm. This could be due to the fact 
that databases are generally expensive to collect and hence rather small.\\ 

\subsubsection{Manhob HCI-Tagging}
The complete Manhob HCI-Tagging database contains 3490 video sequences recorded during 4 sessions,
with constant acquisition conditions. The only difference between the sessions consists in the
task assigned to the participants, and that should not affect the performance of any rPPG algorithm. 
However, there is a great amount of variation in sequence length, since the smallest sequence 
contains 556 frames and the largest 15780, the mean being 1789 frames. Different sets were built based 
on subjects identities, ensuring no overlaps between training and test sets. 
Also, since the sequence duration is usually correlated with the session, 
all sessions corresponding to selected subjects were included in the set. This is done 
to balance the mean sequence duration across the different sets. The training set has been built
using data from 19 subjects and the test set with data of the remaining 10 subjects.
Table~\ref{tab:manhob-sets}  summarizes relevant information about this partitioning.

\begin{table}[!h]
\renewcommand{\arraystretch}{1.3}
\caption{Statistics for training and test sets of the Manhob HCI-Tagging database. The minimum, 
maximum and mean length are given in number of frames.}
\label{tab:manhob-sets}
\centering
\begin{tabular}{|c||c|c|c|c|}
\hline
 Set & \# of sequences & min length & max length & mean length\\
\hline
\hline
Train & 2302 & 556 & 13106 & 1812\\
Test & 1188 & 579 & 15780 & 1745\\
\hline
\end{tabular}
\end{table}

We then applied the different baseline algorithms to the different sets 
of the Manhob HCI-Tagging database. Table~\ref{tab:hci-protocols-opt-train} shows the
Pearson's correlation coefficients obtained when parameters are optimized on the training set.

\begin{table}[!h]
\renewcommand{\arraystretch}{1.3}
\caption{Pearson's correlation coefficient on different sets of the Manhob HCI-Tagging database}
\label{tab:hci-protocols-opt-train}
\centering
\begin{tabular}{|c||c|c|}
\hline
& \textbf{Training set} & \textbf{Test set} \\
\hline
\hline
\textbf{LiCVPR} & \textbf{0.49} & \textbf{0.45}\\
\textbf{CHROM} & 0.15 & 0.14\\
\textbf{2SR} &  0.17 & 0.05\\
\hline
\end{tabular}
\end{table}

As can be seen on Table~\ref{tab:hci-protocols-opt-train}, obtained performance is generally better on the training set than on the test set, as expected. 
Also, it can be observed that performance is not as good as in the previous experiment (Table~\ref{tab:results-baseline-527-cut}). This can be mainly explained by considering
two points. First, the length of the different sequences greatly varies, and some parameters may 
yield to an overall better behaviour if the sequences are of the same duration (i.e. some parameters value
may be best suited to long sequences). As a matter of fact,
optimal parameters on the training set are different than the ones obtained in the previous experiments.
Secondly, and a more obvious reason for this performance drop is that more sequences introduces more variations.\\

\subsubsection{COHFACE}
Since the COHFACE dataset contains two distinct illumination conditions, different
experimental protocols have been devised. For this test however, we only consider 
studio conditions, i.e. when the face is well-lit by the spot (see Figure~\ref{fig:cohface-office}). 
To generate training and test sets, the same approach
as with the Manhob HCI-Tagging database has been applied: 24 subjects were selected to build the training set, and the
16 remaining subjects form the test set. Considering studio conditions only, this partitioning results in 
48 sequences for the training set and 32 for the test set (there are 2 sequences for each subject).\\

\begin{table}[!h]
\renewcommand{\arraystretch}{1.3}
\caption{Pearson's correlation coefficient for the \textit{studio} protocol on different sets of the COHFACE database.}
\label{tab:cohface-protocols-clean}
\centering
\begin{tabular}{|c||c|c|}
\hline
& \textbf{Training set} & \textbf{Test set} \\
\hline
\hline
\textbf{LiCVPR} & -0.16 & -0.61\\
\textbf{CHROM} & \textbf{0.23} & \textbf{0.43}\\
\textbf{2SR} & 0.07 & -0.31\\
\hline
\end{tabular}
\end{table}

Results in Table~\ref{tab:cohface-protocols-clean} should be compared to the ones reported in Table~\ref{tab:results-baseline-527-cut}: sequences are 
approximatively of the same length and recorded in clean illumination conditions.
In this case however, obtained results are much worse for two out of the three studied algorithms. 
On this database, they prove to be incapable at inferring the heart-rate reliably, since their 
Pearson's correlation coefficient is close to zero and even negative in the case of LiCVPR. This result, 
although suprising, is interesting: it emphasizes the sensitivity of these approaches to 
different acquisition conditions. Indeed, the COHFACE database has both lower resolution and framerate.
On the other hand, the performance of the CHROM algorithm remains 
consistent across these different datasets. An interesting point to note is that it actually
performs better on the test set than on the training set. This could mean that its performance
is less sensitive to its parameters, which is a nice feature: it suggests a good
generalization ability to unseen video sequences. 
However, it should be noted that optimal 
parameters for the training set of the COHFACE database are generally not the same as the one used in the 
very first experiment.

\subsection{Evaluation on umatched conditions}
An interesting and realistic scenario is to use rPPG algorithms across a wider range of aquisition 
conditions. Think, for instance, of a usage on a mobile device in different rooms, where the lighting conditions
are not the same. It would be an additional burden to readjust different parameters each time there is a 
change in the environment. Hence it is important to assess the behaviour of the 
different approaches in various settings. Such cases have been emulated with two different experiments presented below.\\

\subsubsection{Cross database}
Although the two considered databases were partly recorded in quite similar conditions (indoor, controlled illumination), 
it is of interest to test the behaviour of the different approaches in a cross-database setting. For this purpose, 
we first tune the algorithms on the training set of the Manhob HCI-Tagging database and test them on the studio test set of
the COHFACE database. For the second experiments, algorithms are optimized on the studio training set of the COHFACE database 
and tested on the test set of the Manhob HCI-Tagging database.

\begin{table}[!h]
\renewcommand{\arraystretch}{1.3}
\caption{Pearson's correlation coefficient in cross-database testing.}
\label{tab:results-xdb}
\centering
\begin{tabular}{|c||c|c|c|}
\hline
& \textbf{LiCVPR} & \textbf{CHROM} & \textbf{2SR}\\
\hline
\hline
\textbf{HCI $\rightarrow$ COHFACE} & -0.20  & \textbf{0.51} & -0.15\\
\hline
\textbf{COHFACE $\rightarrow$ HCI} & \textbf{0.25} & 0.10 & 0.00\\
\hline
\end{tabular}
\end{table}

Results reported in Table~\ref{tab:results-xdb} shows an interesting behaviour. Indeed,
when algorithms are tuned on the Manhob HCI-Tagging database, their performance on the test set of the COHFACE
database is better than when tuned with the same database. However, results for both LiCVPR and 2SR are
still not satisfactory. While such a behaviour may seem surprising, 
it only means that optimal parameters on the training set of Manhob HCI-Tagging also suit well the data
in the COHFACE studio test set. This can be due to the wider variety of sequences to tune parameters, since
the training set of the Manhob HCI-Tagging database contains much more sequences than the training set of
the COHFACE database. In the second experiment however, one can see that performance is 
generally worse than the one reported in the second column of Table~\ref{tab:hci-protocols-opt-train}, as expected.
In this case, the limited amount of sequences in the COHFACE studio training set prevents to find optimal
parameters for the more generic conditions in the test set of the Manhob HCI-Tagging database.
Again, these experiments underline the extreme sensitivity of the different approaches to their respective parameters. Note finally
that the performances of CHROM on both test sets are comparable, suggesting its higher stability to different configurations.\\

\subsubsection{Illumination conditions}
Since the COHFACE database contains two distinct illumination conditions (see Figure~\ref{fig:db-lighting}), one is now able
to test the generalization of rPPG algorithms to different illumination conditions. The number of
recordings under natural lighting is the same as in well-lit conditions, meaning that the natural test set contains
16 subjects and 32 video sequences. To test the behaviour in mismatched conditions, parameters were tuned on the
studio training set, and applied to the natural test set. Performance in unmatched illumination conditions are reported
in Table~\ref{tab:results-illumination-mismatch}.\\

\begin{table}[!h]
\renewcommand{\arraystretch}{1.3}
\caption{Pearson's correlation coefficient on the natural test set of the COHFACE database.}
\label{tab:results-illumination-mismatch}
\centering
\begin{tabular}{|c||c|c|c|}
\hline
& \textbf{LiCVPR} & \textbf{CHROM} & \textbf{2SR}\\
\hline
\hline
\textbf{Studio $\rightarrow$ Natural} & -0.24 & -0.03 & \textbf{0.00}\\
\hline
\end{tabular}
\end{table}

As expected, obtained results generally show a drop in performance as compared to the test set in studio conditions (Table~\ref{tab:cohface-protocols-clean}). 
This denotes again the need for a careful selection of the parameters, and suggest that none of the selected algorithms are suitable
to achieve acceptable performance in realistic scenarios, where the acquisition conditions are unknown \textit{a priori}. \\

\subsection{Importance of the ROI}
In the next experiments, we consider the whole COHFACE dataset, with both lighting conditions. Doing so will 
allow to get insights on the performances in a more generic setting in terms of illumination, corresponding to 
more realistic conditions. Using the full dataset, there are now 96 sequences in the training set, and 64
in the test set. Note that both sets contain sequences recorded under the two different lightings, 
there is no mismatch between training and testing, but rather a wider range of variability. \\

Since all investigated rPPG algorithms first extract skin pixels, it is interesting to see which method is best suited 
to perform this task. As a baseline, the complete face bounding box is used. We then compare the different algorithms
when the skin pixels are either extracted using a mask on the bottom part of the face (see Figure~\ref{fig:block-li}), or using the 
skin color filter \citep{taylor-spie-2014}. Performance using different ROI to extract skin pixels are presented in Table~\ref{tab:pearson-ROI}.\\

\begin{table}[!h]
\renewcommand{\arraystretch}{1.3}
\caption{Pearson's correlation coefficient for different skin color ROI on the test set of the COHFACE database.}
\label{tab:pearson-ROI}
\centering
\begin{tabular}{|c||c|c|c|}
\hline
& \textbf{Face bounding box} & \textbf{Skin pixels} & \textbf{Mask} \\
\hline
\hline
\textbf{LiCVPR} & -0.29 & -0.16 & -0.44\\
\textbf{CHROM} &  \textbf{0.30} & \textbf{0.27} & 0.30 \\
\textbf{2SR} & 0.26 & 0.09 & \textbf{0.65}\\
\hline
\end{tabular}
\end{table}

In this case, the performance of the CHROM algorithm remains stable and comparable to the one obtained on the studio dataset
(see Table~\ref{tab:cohface-protocols-clean}), no matter which ROI is considered. 
This could be explained by the fact that the mean color is used, which should not vary much across
different ROIs. This also shows that when parameters are tuned using conditions known \textit{a priori}, 
this algorithm is able to cope with variability in illumination. 
It is interesting to see that the 2SR algorithm performs better when the mask is considered. Indeed, one
could expect that it would work best when the color subspace of the considered skin pixels are compact in the RGB space, since
it has been designed this way. This result suggests that with this algorithm the ROI has a tremendous influence
on the results, as evidenced by the variation in performance across the different regions.
Finally, one should notice that LiCVPR's algorithm does not achieve acceptable performance no matter the ROI. This shows
that this algorithm is very sensitive to variations in terms of both image resolution and illumination conditions.

\section{Conclusion}
In this contribution, three state-of-the-art rPPG algorithms were selected and evaluated.
For this purpose, a new, publicly available database containing 40 subjects captured under two different
illumination conditions has been introduced. A thorough experimental evaluation of the selected approaches has been conducted 
using different datasets and their associated protocols. Our reproducible research framework allows assessing performance in a principled and 
unbiased way. Obtained results show that only one rPPG algorithm has a stable behaviour, but overall it has been noticed that performance
is highly dependent on a careful optimization of parameters. Conducted experiments also shows that generalization across conditions (i.e. resolution, illumination)
should be of high concern when assessing rPPG approaches. 
The data, the experimental protocols and the implementation of the 
algorithms used in this study have been made available as open source free software\footnote{https://pypi.python.org/pypi/bob.rppg.base}. 
We hope that this will help standardizing the comparison of remote heart rate measurement algorithms and advance the progress in this field.

\section*{Acknowledgment}
This work was supported by the Hasler foundation through the COHFACE project
and by the FP7 European project BEAT (284989). 
The authors would like to thank Sushil Bhattacharjee at Idiap Research Institute for his help with the recordings and for providing the landmarks,
Xiaobai Li at the University of OULU for sharing part of his code, and Weijin Wang at the Technical University of Eindhoven for fruitful discussions.

\bibliographystyle{plain}
\bibliography{biblio}

\end{document}